\documentclass[runningheads]{llncs}
\usepackage{graphicx}

\usepackage{tikz}
\usepackage{comment}
\usepackage{amsmath,amssymb} 
\usepackage{color}
\usepackage{bbold}
\usepackage{enumitem}
\usepackage{cuted}
\usepackage{multirow}
\usepackage{hyperref}
\usepackage[accsupp]{axessibility}  

\usepackage[capitalize]{cleveref}
\crefname{section}{Sec.}{Secs.}
\Crefname{section}{Section}{Sections}
\Crefname{table}{Table}{Tables}
\crefname{table}{Tab.}{Tabs.}

\def\etal{\emph{et al.}}
\newcommand{\thatis}{\textit{i.e.}}
\newcommand{\suchthat}{\textit{s.t.}}
\DeclareMathOperator*{\argmax}{arg\,max}
\DeclareMathOperator*{\argmin}{arg\,min}

\begin{document}
\pagestyle{headings}
\mainmatter

\title{ART-SS: An Adaptive Rejection Technique for Semi-Supervised restoration for adverse weather-affected images} 

\titlerunning{ART-SS}
\author{Rajeev Yasarla\and
Carey E. Priebe\and
Vishal Patel}
\authorrunning{R. Yasarla et al.}
\institute{Johns Hopkins University, Baltimore, MD 21218, USA
\email{\{ryasarl1,cep,vpatel36\}@jhu.edu}}
%

\maketitle

\begin{abstract}
In recent years, convolutional neural network-based single image adverse weather removal methods have achieved significant performance improvements on many benchmark datasets.  However, these methods require large amounts of clean-weather degraded image pairs for training, which is often difficult to obtain in practice.  Although various weather degradation synthesis methods exist in the literature, the use of synthetically generated weather degraded images often results in sub-optimal performance on the real weatherdegraded images due to the domain gap between synthetic and real world images.  To deal with this problem, various semi-supervised restoration (SSR) methods have been proposed for deraining or dehazing which learn to restore clean image using synthetically generated datasets while generalizing better using unlabeled real-world images. The performance of a semi-supervised method is essentially based on the quality of the unlabeled data.  In particular, if the unlabeled data characteristics are very different from that of the labeled data, then the performance of a semi-supervised method degrades significantly.  We theoretically study the effect of unlabeled data on the performance of an SSR method and develop a technique that rejects the unlabeled images that degrade the performance. Extensive experiments and ablation study show that the proposed sample rejection method increases the performance of existing SSR deraining and dehazing methods significantly. Code is available at :\textit{\href{https://github.com/rajeevyasarla/ART-SS}{https://github.com/rajeevyasarla/ART-SS}}
\end{abstract}

\section{Introduction}

	\label{sec:intro}
	
	
	Images captured in weather degradations like rain or fog  conditions are of poor quality, leading to a loss of situational awareness and a general decrease in usefulness.  Hence, it is very important to compensate for the visual degradation in images caused by these weather degradations.  Additionally, such weather degraded images also reduce the performance of down-stream computer vision tasks such as detection, segmentation and recognition \cite{li2018benchmarking,chen2018encoder,ren2015faster}.  
	The main objective Single image restoration (SIR) of weather degraded image, is to restore the clean image $y$, given a weather degraded image $x$, in-order to improve performance of such down-stream tasks. Extensive research on methods to remove such weather degradation effects like rain and haze.
	
	\begin{figure}[h!]
		\centering
		\includegraphics[width=\linewidth]{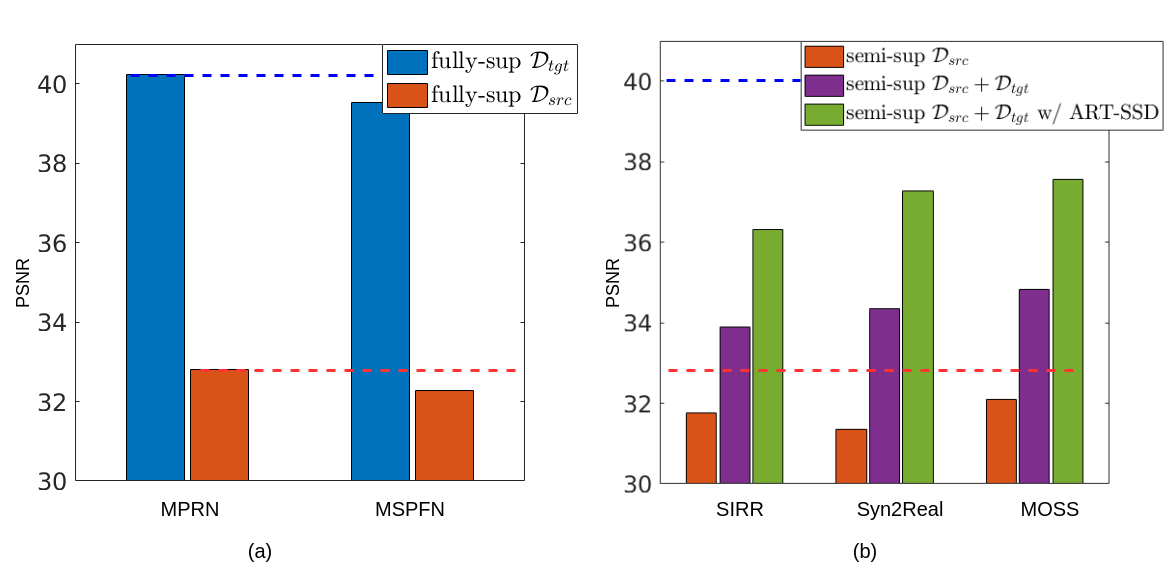} 
		\vskip -10pt 
		\caption{Cross-domain deraining experiment where Rain800~\cite{zhang2019image}  is used as the synthetic source dataset $\mathcal{D}_{src}$, and SPA-data~\cite{wang2019spatial} is used as the real rain target dataset $\mathcal{D}_{tgt}$.  Here, MPRN~\cite{zamir2021multi} and MSPFN~\cite{jiang2020multi} are fully-supervised methods and SIRR~\cite{wei2019semi}, Syn2Real~\cite{Yasarla_2020_CVPR} and MOSS~\cite{Huang_2021_CVPR} are semi-supervised methods.  Fully-supervised methods are supervised using the corresponding labeled clean images and semi-supervised images are trained using a labeled source data $\mathcal{D}_{src}$ and an unlabeled target data $\mathcal{D}_{tgt}$.  (a)  Blue and Red bars show the target only and source only performance of MSPFN and MPRN, respectively. We can see a drop in performance of supervised methods when they are trained on $\mathcal{D}_{src}$ and tested on $\mathcal{D}_{tgt}$. (b)  Semi-supervised methods such as SIRR, Syn2Real and MOSS perform better than the supervised methods by leveraging the information from unlabeled images.  However, with the help of ART-SS, we are able improve the performance of semi-supervised restoration(SSR) methods even further by rejecting the unlabeled images that are not helpful in semi-supervision.}
		\label{fig:bar_graph}
		\vspace{-2.5em}
	\end{figure}
	
	In recent years, various convolutional neural network-based methods have been proposed for deraining\cite{fu2017removing,yang2017deep,wang2019spatial,li2018recurrent,ren2019progressive,fu2021rain,zhou2021image,wang2021rain,lin2020rain}, dehazing\cite{dong2020multi,li2020deep,wu2021contrastive,zhang2018density,zhang2019joint,zhang2021hierarchical}.  These fully-supervised methods require  large amounts of clean-weather degraded image pairs for training.  Since collecting real world weather degraded-clean image pairs of data is difficult, most existing supervised methods rely on synthetically generated data to train the network.  However, the use of synthetically generated weather degraded images often results in sub-optimal performance on the real world images due to the domain difference between synthetic and real world images.  For example when we consider deraining task, this can be clearly seen in Fig.~\ref{fig:bar_graph} (a), where we train two fully-supervised SID networks, MPRN~\cite{zamir2021multi} and MSPFN~\cite{jiang2020multi}, on a synthetic source dataset $\mathcal{D}_{src}$ from Rain800~\cite{zhang2019image} and test them on a real rain target dataset $\mathcal{D}_{src}$ from SPA-data \cite{wang2019spatial}. From Fig.~\ref{fig:bar_graph} (a), we can observe that the performance of fully-supervised methods degrades significantly when trained on Rain800~\cite{zhang2019image} and tested on SPA-data\cite{wang2019spatial} compared to target only performance which corresponds to the case where the methods are trained and tested on SPA-data \cite{wang2019spatial}.

	To address this domain gap between source and target datasets, Wei \etal\cite{wei2019semi} initially attempted to address the semi-supervised deraining task by leveraging the rain information in unlabeled target dataset while training the network. In proposed method, the authors model rain residuals by imposing a likelihood term on Gaussian Mixture Models (GMMs) for both labeled and unlabeled datasets, and minimize minimize the Kullback-Leibler (KL)
	divergence between the obtained GMMs of labeled and unlabeled images to enforce the consistency that distribution of labeled rainy data should be close to that of the unlabeled data. Later following this approach,
	Yasarla \etal~\cite{Yasarla_2020_CVPR} proposed a non-parametric model for semi-supervised deraining where they project labeled and unlabeled rainy images to a latent space and formulate a joint Gaussian distribution to  generate pseudo labels for the unlabeled images. Recently, Huang~\etal~\cite{Huang_2021_CVPR} proposed a memory-based encoder-decoder network where the memory module learns rain information from synthetic and real rainy images in a self-supervised manner using Exponential Moving Average (EMA) updates. On the other hand, to address the semi-supervised dehazing Li~\etal\cite{li2019semi} proposed to leverage hazy information from unlabeled images using dark channel priors based gradient updates while training the network. Later, Shao~\etal\cite{shao2020domain} proposed a bi-directional translations method that minimizes the gap between synthetic and real hazy domains using adversarial loss and dark channel priors.
	
	One major drawback of these semi-supervised restoration(SSR) techniques is that they don't account for the effect of unlabeled images on the overall semi-supervised performance.  Not all images in the unlabeled target dataset are useful in improving the SSR performance.  If the unlabeled image characteristics are very different from that of in the source data, then there is a good chance that SSR performance will converge to unsupervised deraining performance instead of converging towards fully-supervised. We explore theoretical evidence for this behavior, and also conduct cross-domain experiments to empirically show that unlabeled observations which are different from the labeled source images might not be beneficial in improving the SSR performance. 

    In particular, we theoretically understand why a few unlabeled observations might have an adverse effect on the performance of an SSR method, and propose a novel technique called \underline{a}daptive \underline{r}ejection \underline{t}echnique for \underline{s}emi-\underline{s}upervision (ART-SS), that selects unlabeled observations which are useful in improving the SSD performance.  In Syn2real~\cite{Yasarla_2020_CVPR} and MOSS~\cite{Huang_2021_CVPR}, authors project the unlabeled and labeled images to a latent space and express latent vector of each image using either latent basis vector representations or labeled latent-space vectors. Additionally, these works perform supervision at the defined latent space level in unlabeled trainning phase of semi-supervised training. Following these works, we use the latent representation of the labeled or unlabeled images and compute similarity index ($\psi$) for each image that indicates how similar is the given image to the labeled images. Note as the unlabeled images can be easy or hard samples, and network might produce errors in computing the latent representations, thus we compute the variance $\sigma$ (aleotoric uncertainty~\cite{kendall2017uncertainties}) that indicates a measure of how confident the network is about computing the latent representation. Hence, we use proposed theorem and corollaries in our theoretical study, and come up with a novel selection criterion for ART-SS method using $\psi$ and $\sigma$ measures, to decide whether the given unlabeled image is helpful for improving the SSR performance. Note, using variance $\sigma$ (aleotoric uncertainty) makes the ART-SS method robust to error in the networks latent-space representations. In this way given unlabeled image, we compute $\psi$ and $\sigma$ measures for the unlabeled image, and using the criterion to decide whether the unlabeled image is similar or dis-similar(\thatis~might have adverse affect on SSR performance)  to source domain, and can be used for updating the weights of a SSR method or not.  For example using proposed ART-SS, we are able to significantly boost the performance of existing SSR deraining methods~\cite{wei2019semi,Yasarla_2020_CVPR,Huang_2021_CVPR} (see Fig~\ref{fig:bar_graph}(b)). 
		
	In summary, this paper makes the following contributions:
	\begin{itemize}[noitemsep,nolistsep]
		\item We theoretically study how unlabeled images can affect the performance of an semi-supervised restoration(SSR) method.
		\item We propose a novel rejection technique, called ART-SS, to select images that are useful in improving the SSR performance.  
		\item Extensive cross-domain experiments and ablation study are conducted to show the significance of the proposed method.  In particular, our simple rejection technique is shown to  boost the performance of the existing deraining \cite{wei2019semi,Yasarla_2020_CVPR,Huang_2021_CVPR}, and dehazing\cite{li2019semi,shao2020domain} methods.
	\end{itemize}
	
	\section{Related work}
	Various single image restoration methods have been proposed for adverse weather removal problems like deraining\cite{wang2020model,wang2019erl,hu2019depth,li2019heavy,li2020all,wang2020rethinking,kang2011automatic,huang2013self,luo2015removing,li2016rain}, dehazing\cite{he2010single,ren2016single,ren2018gated,berman2016non,fattal2014dehazing,li2018single}. Here, we mainly focus on SSR tasks in deraining, and dehazing.
	
	\noindent{\textbf{Deraining.}} Jiang~\etal~\cite{jiang2020multi} proposed a fusion network called MSPFN that fuses hierarchical deep features in a progressive manner for deraining. Zamir~\etal\cite{zamir2021multi} proposed a multi-stage architecture that incorporates the information exchange between different stages in retrieving the derained image. As these methods are trained on synthetic rainy-clean pairs, these methods might obtain sub-optimal performances when tested on real rainy images since there is domain gap between synthetic and real rainy images. To this end, semi-supervised approaches have been proposed Wei~\etal~\cite{wei2019semi} (GMM based), Yasarla~\etal\cite{Yasarla_2020_CVPR} (Gaussian process based pesudo-GT generation), Huang~\etal\cite{Huang_2021_CVPR} (mean student teacher learning) to address the domain gap between synthetic and real rainy images to improve SSR performance.\\
	\noindent{\textbf{Dehazing.}} Ren~\etal~\cite{ren2018gated} pre-processed a hazy image to generate multiple input images, hence introducing color distortions to perform dehazing. Qu~\etal~\cite{qu2019enhanced} proposed an enhanced image-to-image translation based dehazing method trained using adversarial loss. Although these acheive better performance on synthetic hazy images, might fail to restore high quality clean image given real hazy image. To this end, \cite{li2019semi,shao2020domain} proposed semi-supervised dehazing approaches using dark channel priors, and total-variation loss to reduce domain gap between synthetic and real hazy images. 
	
	These SSR methods don't account for the effect of unlabeled image on semi-supervised performance and might suffer to obtain optimal semi-supervised performance gains. Inspired by Yang and Priebe~\cite{yang2011effect}, we theoretically study the effect of unlabeled data on the SSR performance and develop a rejection technique that rejects  unlabeled images which are not beneficial in improving the SSR performance.

	\section{Preliminaries}
	In this section, we define notations and key concepts regarding specified and  misspecified models and present a semi-supervised degradation theorem.

	\subsection{Model and notations}
	Given, a weather-degraded image $x$, our objective is to obtain a restored image $\hat{y} = f(x)$, where $f(.)$ is a function with parameters $\theta$ that performs the restoration(deraining or dehazing) task.  This function can be any deep learning-based model or a GMM-based model.  Let us denote  the collection of all possible restoration functions $\{f(.)\}$ in the parametric model $\mathcal{F}$ whose parameters  expressed as $\Theta_{\mathcal{F}}$ by a dashed circle.  Let $f_{opt}(.)$  denote the best possible deraining function in $\mathcal{F}$, i.e.,
	\setlength{\belowdisplayskip}{0pt} \setlength{\belowdisplayshortskip}{0pt}
    \setlength{\abovedisplayskip}{0pt} \setlength{\abovedisplayshortskip}{0pt}
	\begin{equation}
	f_{opt} = \argmin_{f \in \mathcal{F}} L(f),
	\end{equation}
	where  $L(.)$ is used to denote the error for the function $f(.)$ in the restoration task. Bayes error (the lowest possible error that can be achieved and is the same as irreducible error)  is expressed as $L^*$ and the corresponding function as $f^*$.  Let $\hat{f}$ be the learned restoration function with  parameters  $\hat{\theta}$. The model bias is measured by $L(f_{opt}) -L^*$ and the estimation error is $L(\hat{f}) - L(f_{opt})$. Now let us define the limits depending on whether we are learning the restoration task in supervised fashion, ($L_{sup}^*,\: f_{sup}^*,\: \theta_{sup}^*$) or  unsupervised fashion ($L_{unsup}^*,\: f_{unsup}^*,\: \theta_{unsup}^*$). We denote error for the fully-supervised method using labeled data as $L_\ell$, and semi-supervised method using labeled and unlabeled as $L_{\ell+u}$.

	We denote the labeled source dataset as $\mathcal{D}_{src}$, and the target unlabeled dataset as $\mathcal{D}_{tgt}$. Following Syn2real~\cite{Yasarla_2020_CVPR} and MOSS~\cite{Huang_2021_CVPR}, we project the labeled and unlabeled datasets onto a latent space which is defined as the output of an encoder.  That is,  every image $x^l_i \in \mathcal{D}_{src}$ is passed through the encoder network to obtain $z^l_i = g(x^l_i)$.  Similarly, $z^u_i = g(x^u_i)$ is obtained for every image $x^u_i \in \mathcal{D}_{tgt}$. Note that encoder and decoder of a restoration network are represented using functions $g(.)$ and $h(.)$, with corresponding parameters $\theta^{enc}$ and $\theta^{dec}$, respectively.  For the sake of simplicity,  let us assume that all the labeled latent vectors,  $Z_{src}$ =$\{z^l_i\}$ can be spanned by a set of vectors $\{\{s_i\}^{M_l}_{i=1},\{c_j\}_{j=1}^{M_c}\}$, \thatis $z^l_i \in span(\{\{s_i\}^{M_l}_{i=1},\{c_j\}_{j=1}^{M_c}\})$ or $z^l_i~=~\sum_{v_i \in \{\{s_i\}^{M_l}_{i=1},\{c_j\}_{j=1}^{M_c}\}} \alpha_i v_i $ and  we represent this vector space as $\mathcal{V}_{src}$ . Similarly,  all unlabeled latent vectors, $Z_{tgt}$ =$\{z^u_i\}$ can be spanned by a set vectors $\{\{t_i\}^{M_u}_{i=1},\{c_j\}_{j=1}^{M_c}\}$, \thatis $z^u_i \in span(\{\{t_i\}^{M_u}_{i=1},\{c_j\}_{j=1}^{M_c}\})$ or $z^u_i~=~\sum_{v_i \in \{\{t_i\}^{M_u}_{i=1},\{c_j\}_{j=1}^{M_c}\}} \alpha_i v_i $ and we represent this vector space as $\mathcal{V}_{tgt}$.  Note that we have assumed that the labeled vector space $\mathcal{V}_{src}$  and the unlabeled vector space $\mathcal{V}_{tgt}$, have common basis vectors $V_c = \{c_j\}_{j=1}^{M_c}$ (because of similarities between labeled and unlabeled images).  In addition,  these vector spaces $\mathcal{V}_{src}$ and $\mathcal{V}_{tgt}$ have different basis vectors $V_s = \{\{s_i\}^{M_l}_{i=1}$ and $V_t = \{\{t_i\}^{M_u}_{i=1}$ respectively (this is due to differences  or domain gap between the  labeled and unlabeled weather-degraded images). 
	
	\subsection{Correct parametric model (specified)}
	If $f^*\in \mathcal{F}$, then the model bias is $0$, \thatis, $ {L}(f_{opt}) - {L}^* = 0
	$. The estimation error is the only thing that contributes to the regression error of the weather removal task. In other words, $V_s = \{\{s_i\}^{M_l}_{i=1} = \emptyset$ and $V_t = \{\{t_i\}^{M_u}_{i=1}  = \emptyset$, where $ \emptyset$ denotes the  empty set.  In other words,  model $\mathcal{F}$ is good enough in learning the weather removal function $f(.)$  that minimizes the difference between  labeled and unlabeled weather-degraded images. In the parametric setting, if we use Mean Squared Error (MSE) on the parameter space $\Theta_{\mathcal{F}}$, then we have 
	\setlength{\belowdisplayskip}{0pt} \setlength{\belowdisplayshortskip}{0pt}
    \setlength{\abovedisplayskip}{0pt} \setlength{\abovedisplayshortskip}{0pt}
	\begin{equation} \label{eqn:model_mse}
	MSE(\hat{\theta}) = \mathbf{E}\left[\left(\hat{\theta}-\theta\right)^2\right] =  \left(\mathbf{E}[\hat{\theta}]-\theta \right)^2 + Var(\hat{\theta}).
	\end{equation}
	The term $\left(\mathbf{E}[\hat{\theta}]-\theta \right)^2 $ is a form of bias. In a correct parametric model, fully-supervised and semi-supervised deep learning models converge to the same parameter value $\theta^*$. In other words, both fully-supervised error and semi-supervised error tends to $L^*$, \thatis\; $L_l\rightarrow L^*$, and $L_{l+u}\rightarrow L^*$, as $N_\ell\rightarrow\infty$ and $\frac{N_\ell}{N_u}\rightarrow 0$, where $N_\ell$ and $N_u$ represent the number of labeled and unlabeled images.

	\subsection{Incorrect parametric model (misspecified)}
	If $f^*\notin \mathcal{F}$, then  $ {L}(f_{opt}) - {L}^* > 0$. In this case we change the training set from $\mathcal{D}_{src}$ (labeled) to $\mathcal{D}_{src} +\mathcal{D}_{tgt} $.  However, this will only change the estimation error (in Eq.~\ref{eqn:model_mse}). Adding unlabeled observations reduces the estimation variance.  Nonetheless, fully-supervised and semi-supervised deep learning models may converge to different parameter values. In other words, model $\mathcal{F}$ isn't good enough to learn a deraining function $f(.)$  that minimizes labeled and unlabeled weather-degraded images.  There exists domain gap between latent labeled and unlabeled vectors, and  $V_s \neq \emptyset$ and $V_t  \neq \emptyset$. Given a fixed number of weather-degraded images in the labeled training set, increasing the unlabeled observations may cause a larger estimation bias, \thatis
	\begin{equation}
	\left(\mathbf{E}[\hat{\theta}_{l+u}]-\theta \right)^2 > \left(\mathbf{E}[\hat{\theta}_l]-\theta \right)^2,
	\end{equation}
	where $\hat{\theta}_l$ and $\theta_{l+u}$ are the parameters of fully-supervised and semi-supervised methods. In this case, semi-supervised performance would be degraded if the increase in  estimation bias is more significant than the decrease in the estimation variance. 
	
	\subsection{Semi-supervised degradation theorem}
	Before discussing about a lemma and a theorem for the degradation in semi-supervised(SS) performance, we construct a few idealizations that are required. Let $L(\hat{f})$ be the regression error of a learned restoration function $\hat{f}$($\in\mathcal{F}$), and $KL(f_{\theta^*_{sup}}||\hat{f}) $ be the Kullback-Leibler divergence between fully-supervised limit density and the estimated $\hat{f}$. Here, we assume that $L_{sup}^*,\: f_{sup}^*,\: \theta_{sup}^*$ is the best possible fully-supervised parameters that can be learned given the model $\mathcal{F}$. Similarly, $L_{unsup}^*,\: f_{unsup}^*,\: \theta_{unsup}^*$ denote the best possible unsupervised parameters that can be learned given the model $\mathcal{F}$. 
	
	\noindent{\textbf{Lemma.}} For any fixed finite $N_\ell$ or $N_\ell \rightarrow \infty$, as $\frac{N_\ell}{N_u}\rightarrow 0$, the limit of the maxima of semi-supervised likelihood function reaches the unsupervised limit  $\theta_{unsup}^*$. That is, let $\hat{\theta}_{l+u}$ denote the parameters of a learned SS method when the  number of labeled and unlabeled images are $N_\ell$ and $N_u$, then as $\frac{N_\ell}{N_u}\rightarrow 0$,
	\begin{equation}
	\left\{\hat{\theta}_{(l+u)}\right\}_{u} \stackrel{p}{\longrightarrow} \theta_{u n s u p}^{*} \quad \forall N_\ell
	\end{equation}
	\noindent{\textbf{Proof.}} In semi-supervised learning the samples are drawn from a collection $\mathcal{D}_{src}+\mathcal{D}_{tgt}$ which implies that the probability of drawn realization being labeled image is $\lambda=\frac{N_\ell}{N_\ell+N_u}$, and being unlabeled image is $1-\lambda=\frac{N_u}{N_\ell+N_u}$. The optimization involved in learning the parameters $\hat{\theta}$, is as follows,
	\begin{equation}\label{eqn:expect_opt}
		\argmax_{\theta} \left(\lambda \mathbf{E}_{f(x,y)}[\log f(x,y \mid \theta)]+(1-\lambda) \mathbf{E}_{f(x, y)}[\log f(x \mid \theta)]\right),
	\end{equation}
	which is a convex combination of the fully-supervised and unsupervised expected log-likelihood functions. For an arbitrary finite value of $N_\ell$, as $\frac{N_\ell}{N_u}\rightarrow 0$, $\lambda\:=\frac{N_\ell}{N_\ell+N_u}\rightarrow0$, indicating the above optimization in $\hat{\theta}$, maximizes $\mathbf{E}_{f(x, y)}[\log f(x \mid \theta)]$, which by definition is $\theta_{u n s u p}^{*} $,. Thus the learned semi-supervised parameters, $\left\{\hat{\theta}_{(l+u)}\right\}_{u} \stackrel{p}{\longrightarrow} \theta_{u n s u p}^{*} \quad \forall N_\ell$.
	
	\noindent{\textbf{Theorem.}} 
	If $L(f_{\theta^*_{sup}})<L(f_{\theta^*_{unsup}})$, then for fixed $N_\ell$ or $N_\ell \rightarrow \infty$, as $\frac{N_\ell}{N_u}\rightarrow 0$, 
	\begin{equation*}
		\mathbb{1}\left\{L\left(f_{\hat{\theta}_{\ell}}\right)<L\left(f_{\hat{\theta}_{(\ell+u)}}\right)\right\}
		-\mathbb{1}\left\{K L\left(f_{\theta_{s u p}^{*}} \| f_{\hat{\theta}_{\ell}}\right)<K L\left(f_{\theta_{s u p}^{*}} \| f_{\hat{\theta}_{\ell+u}}\right)\right\} \stackrel{p}{\longrightarrow} 0
	\end{equation*}
	and we have
	\begin{equation*}
	\resizebox{0.95\hsize}{!}{$
		\lim _{N_\ell \rightarrow \infty,\:\frac{N_\ell}{N_u}\rightarrow 0} P\left\{L\left(f_{\hat{\theta}_{\ell}}\right)<L\left(f_{\hat{\theta}_{(\ell+u)}}\right)\right\} =\lim _{N_\ell \rightarrow \infty} P\left\{K L\left(f_{\theta_{s u p}^{*}} \| f_{\hat{\theta_{\ell}}}\right)<K L\left(f_{\theta_{s u p}^{\mathbf{x}}} \| f_{\theta_{u n s u p}^{*}}\right)\right\}$}.
	\end{equation*}
	\noindent{\textbf{Proof.}} Please refer to the supplementary document for the  proof. We use these theoretical results, and come-up with the following corollaries.
	
	\noindent{\textbf{Corollary 1.}} If $L(f_{\theta^*_{sup}}) <L(f_{\theta^*_{unsup}})$, then for the misspecified model, $\exists \ell $, \textit{s.t.}
	\begin{equation*}
	\lim _{N_u \rightarrow \infty} P\left\{L\left(f_{\hat{\theta}_{\ell}}\right)<L\left(f_{\hat{\theta}_{(\ell+u)}}\right)\right\}>0.
	\end{equation*}
	\thatis~ semi-supervised task yields degradation with positive probability as $N_u~\rightarrow~\infty$
	
	\noindent{\textbf{Proof.}}  Please refer to the supplementary document for proof.
	
	\noindent{\textbf{Corollary 2.}} If for a subset of unlabeled images $\mathcal{T}_1\subset \mathcal{D}_{tgt}$, $\exists\:\text{very small}\: \epsilon>0$, \suchthat $|L(f_{\theta^*_{sup}}) -L(f_{\theta^*_{unsup,\mathcal{T}_1}})|<\epsilon$, then 
	\begin{equation}
	\left\{\hat{\theta}_{(l+u)}\right\}_{u\in \mathcal{T}_1} \stackrel{p}{\longrightarrow} \theta_{s u p}^{*} \quad \forall N_\ell.
	\end{equation}
	In other words, model $\mathcal{F}$ behaves nearly like a specified model on the labeled images in $\mathcal{D}_{src}$, and unlabeled images in $\mathcal{T}_1$, since the unlabeled images from subset $\mathcal{T}_1$ are very similar to the labeled images in $\mathcal{D}_{src}$.
	
	\noindent{\textbf{Proof.}} From Eq.~\ref{eqn:expect_opt}, the optimization for learning parameters $\hat{\theta}$ is\\ $
	\argmax_{\theta} \left(\lambda \mathbf{E}_{sup} +(1-\lambda) \mathbf{E}_{unsup} \right)$, where $\mathbf{E}_{sup} = \mathbf{E}_{f(x,y)}[\log f(x,y \mid \theta)]$, and $\mathbf{E}_{unsup} =  \mathbf{E}_{f(x, y)}[\log f(x \mid \theta)]$. We rewrite, $\mathbf{E}_{unsup} = \mathbf{E}_{\mathcal{T}_1} + \mathbf{E}_{\mathcal{D}_{tgt}-\mathcal{T}_1} $, where $\mathbf{E}_{\mathcal{T}_1} = \mathbf{E}_{f(x, y)}[\log f(x \mid \theta, x\in\mathcal{T}_1)]$ and $\mathbf{E}_{\mathcal{D}_{tgt}-\mathcal{T}_1} = \mathbf{E}_{f(x, y)}[\log f(x \mid \theta, x\in \mathcal{D}_{tgt}-\mathcal{T}_1)]$. Thus optimization for learning parameters $\hat{\theta}$  is,
	\begin{equation*}
	\argmax_{\theta} \left(\lambda \mathbf{E}_{sup} +(1-\lambda) (\mathbf{E}_{\mathcal{T}_1} + \mathbf{E}_{\mathcal{D}_{tgt}-\mathcal{T}_1} ) \right)
	\end{equation*}
	if we learn parameter $\hat{\theta}$ for a semi-supervision task using only  $\mathcal{T}_1$ and $\mathcal{D}_{src}$.  That is, rejecting unlabeled observations from  $\mathcal{D}_{tgt}-\mathcal{T}_1$ while learning $\hat{\theta}$. Thus the resultant optimization for learning parameters $\hat{\theta}$  is
	\begin{equation}
		\argmax_{\theta} \left(\lambda \mathbf{E}_{sup} +(1-\lambda) \mathbf{E}_{\mathcal{T}_1}  \right) \approx \argmax_{\theta} \left(\lambda \mathbf{E}_{sup} +(1-\lambda) \mathbf{E}_{sup}  \right). 
	\end{equation}
	Since $|L(f_{\theta^*_{sup}}) -L(f_{\theta^*_{unsup,\mathcal{T}_1}})|<\epsilon$, or unlabeled images from $\mathcal{T}_1$ are  similar  to the labeled images $\mathcal{D}_{src}$, and have similar error, we approximate the optimization for $\hat{\theta}$  to $\mathbf{E}_{sup} = \mathbf{E}_{f(x,y)}[\log f(x,y \mid \theta)]$. Thus, $\left\{\hat{\theta}_{(l+u)}\right\}_{u\in \mathcal{T}_1} \stackrel{p}{\longrightarrow} \theta_{s u p}^{*}$.
	
	The key takeaway from the above theorem and Corollaries is that if the SSR is misspecified, then as increasing the unlabeled images might degraded the SSR performance. In such cases, to boost the SSR performance we can create subset of unlabeled images($\mathcal{T}_1$) by rejecting the unlabeled images that are adversely effecting SSR performance. By doing this SSR will nearly act like a specified model  on $\mathcal{T}_1$ and $\mathcal{D}_{src}$, and semi-supervised performance of SSR tends towards fully-supervised performance.
	
	\vspace{-2em}
	\begin{figure}[h!]
		\centering
		\includegraphics[width=\linewidth]{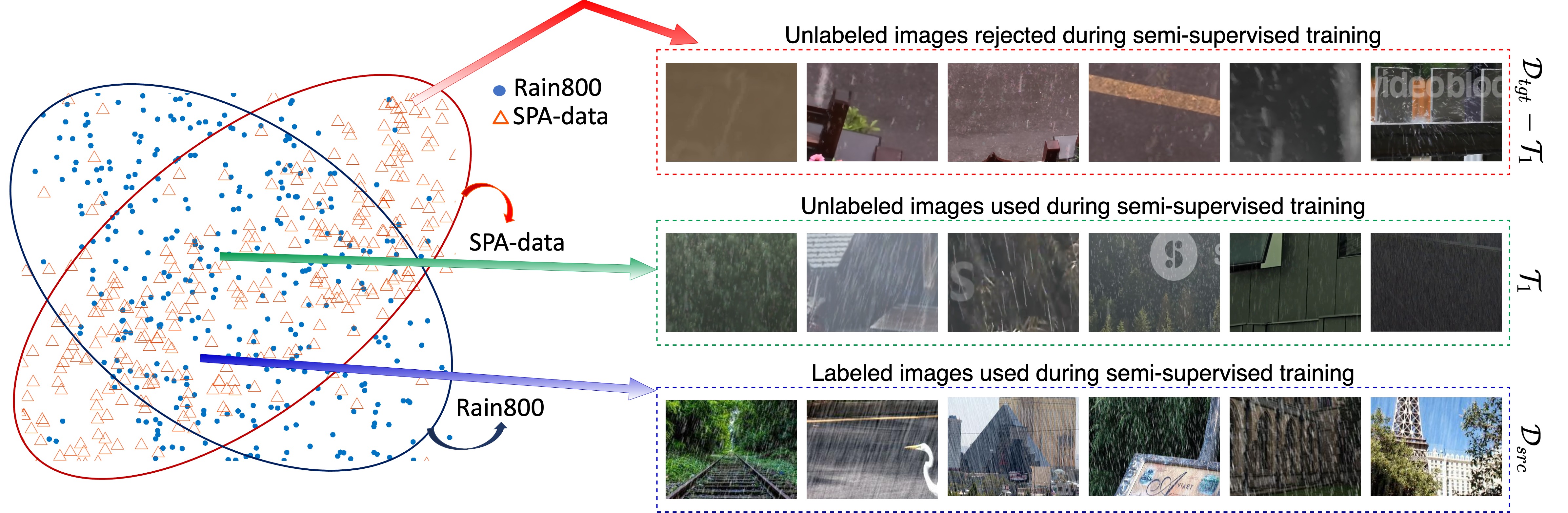} 
		\vskip -10pt 
		\caption{t-SNE plot of Syn2Real\cite{Yasarla_2020_CVPR} for cross-domain experiment with $\mathcal{D}_{src}=\text{Rain800}$ and $\mathcal{D}_{tgt}=\text{SPA-data}$. Here, we can see SSR is misspecified since $V_s = \{s_i\}_{i=1}^{M_l} \neq \emptyset$ and $V_t = \{t_i\}_{i=1}^{M_u} \neq \emptyset$. In order to boost Syn2Real performance we need to create $\mathcal{T}_1$, and train using $\mathcal{D}_{src}+\mathcal{T}_1$, \thatis~ not using unlabeled images from $\mathcal{D}_{tgt}-\mathcal{T}_1$. Rain streaks in $\mathcal{D}_{tgt}-\mathcal{T}_1$ are very different, for example in images 1,3 and 6 rain streaks are curved or look like irregular patches or rain streaks pointing in all directions. On the other hand unlabeled images from $\mathcal{T}_1$ are similar to $\mathcal{D}_{src}$}
		\label{fig:t-SNNE}
	\end{figure}
	\vspace{-3em}
	\section{Proposed method}

	Let an SSR method $f_{\hat{\theta}} \in \mathcal{F}$,  leveraging weather information from unlabeled and labeled images to learn the parameters $\hat{\theta}$. As discussed in the previous section, if the model $\mathcal{F}$ is missepcified, then a domain gap can exist between the unlabeled and labeled images.  In other words, the projected latent labeled and unlabeled vectors can have some different basis vectors, implying  $V_s = \{s_i\}_{i=1}^{M_l} \neq \emptyset$ and $V_t = \{t_i\}_{i=1}^{M_u} \neq \emptyset$. For example in the Fig.~\ref{fig:t-SNNE} t-SNE plot of Syn2Real\cite{Yasarla_2020_CVPR} for cross-domain experiment with $\mathcal{D}_{src}=\text{Rain800}$ and $\mathcal{D}_{tgt}=\text{SPA-data}$, we can see some unlabeled images are similar or close to labeled images and others are not. In such cases we can use Corollary 2 and approximate the model $\mathcal{F}$ as a specified model by training on the labeled dataset $\mathcal{D}_{src}$ and on a subset of unlabeled images ${\mathcal{T}_1}\subset \mathcal{D}_{tgt}$. In this way, we make the model $\mathcal{F}$ behave as nearly specified model on $\mathcal{D}_{src}+{\mathcal{T}_1}$, and can boost the performance of a SSR method,~\thatis~ training SSR on $\mathcal{D}_{src}+{\mathcal{T}_1}$ improves SSR performance towards fully-supervised performance. To this end, we propose ART-SS that rejects unlabeled images that are not similar to labeled images or adversely effecting the SSR performance while training the SSR method.
	\vspace{-1.5em}
	\begin{figure}[h!]
		\centering
		\includegraphics[width=0.6\linewidth]{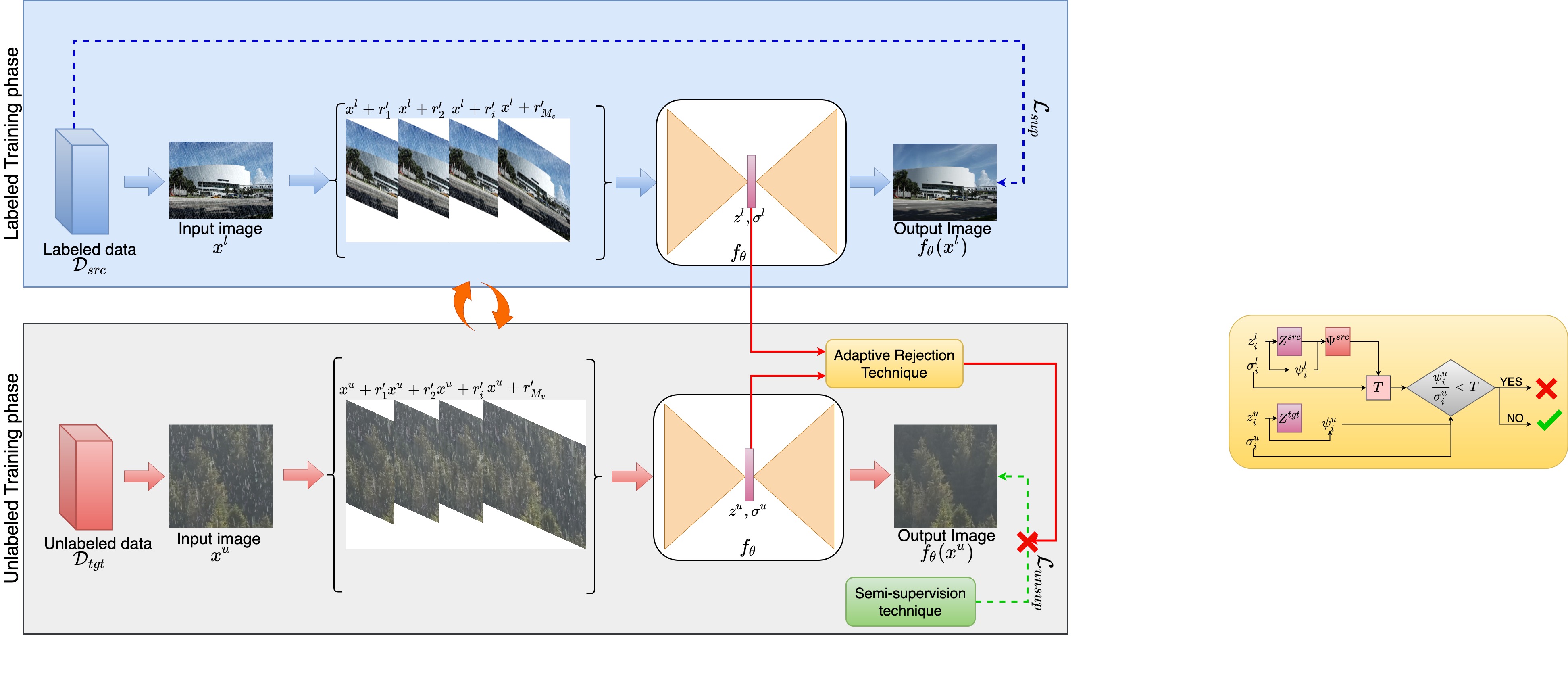} 
		\includegraphics[width=0.38\linewidth]{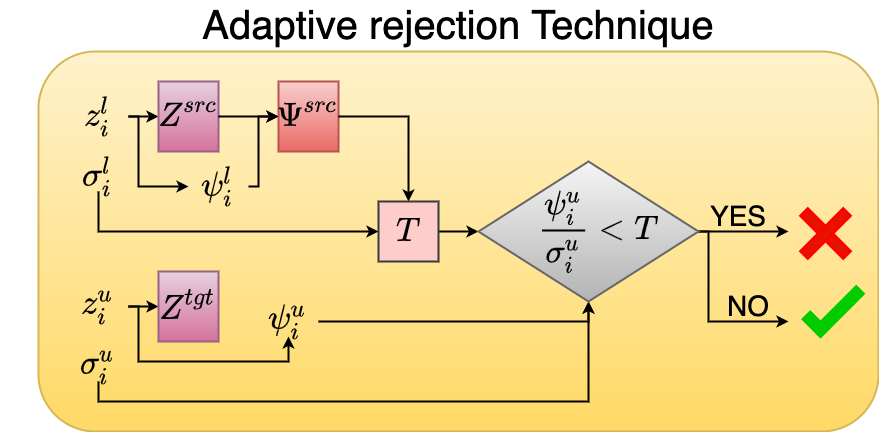} 
		\vskip -5pt 
		\caption{\textbf{Overview of the proposed adaptive rejection technique}. In our rejection, labeled and unlabeled images are projected to the latent space to obtain $z^l \in Z_{src}$ and $z^u \in Z_{tgt}$. Given $z_u,\:Z_{tgt},\:Z_{src}$, we use a rejection module to decide whether to update the network weights  $f_\theta$ using $x_u$ or not. Note that the semi-supervised technique in this figure can be one of \cite{wei2019semi,Yasarla_2020_CVPR,Huang_2021_CVPR}. Here ``tick" in green means perform SSD using $x^u_i$ unlabeled image, and ``red x" in means don't perform SSR using  $x^u_i$.}
		\label{fig:Method}
	\end{figure}
	\vspace{-3.5em}
	\begin{figure}[h!]
		\centering
		\includegraphics[width=1.0\linewidth]{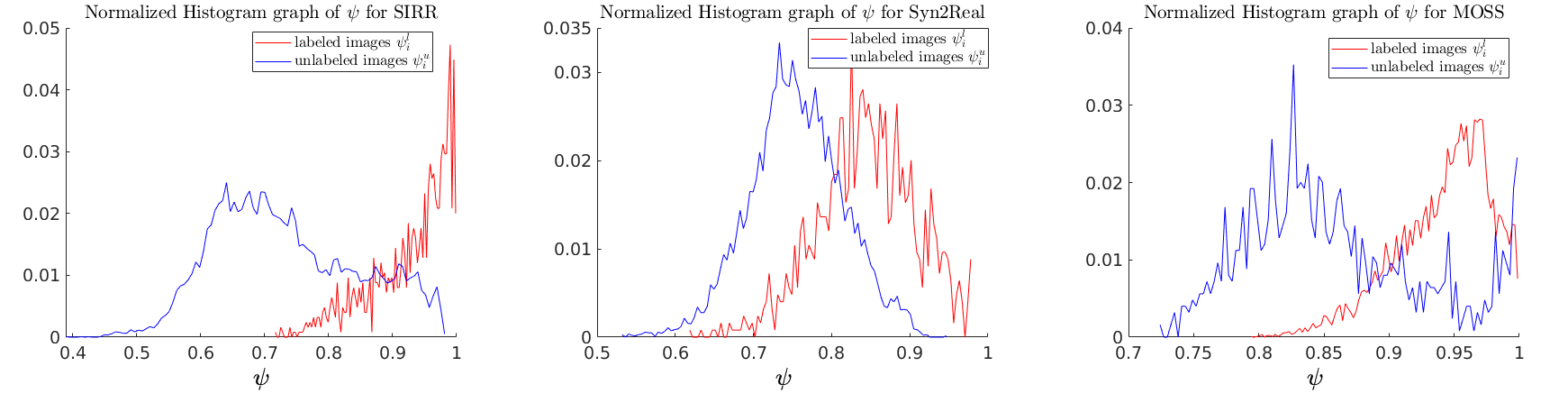} 
		\vskip -10pt 
		\caption{Normalized histogram graphs of $\psi$ for a cross-domain experiment where $\mathcal{D}_{src}$ is Rain800~\cite{zhang2019image} and  $\mathcal{D}_{tgt}$ is SPA-data\cite{wang2019spatial}. Three graphs correspond to three different SSD methods ~\cite{wei2019semi,Yasarla_2020_CVPR,Huang_2021_CVPR}.}
		\label{fig:psi_graph}
	\end{figure}
    \vspace{-2em}
	\subsection{Adaptive rejection technique}\label{sec:reject_tech}
	Fig~\ref{fig:Method} gives an overview of the proposed method where we introduce a rejection module in order to carefully reject the unlabeled observations in $\mathcal{D}_{tgt}$ that are effecting the performance of SSD methods. In our ART-SS method, we project labeled and unlabeled images from $\mathcal{D}_{src}$ and $\mathcal{D}_{tgt}$, to obtain latent vectors $Z_{src}$ and $Z_{tgt}$ respectively. Note \cite{Yasarla_2020_CVPR,Huang_2021_CVPR} express the latent vectors of labeled and unlabeled images using either fixed number of basis latent vectors~\cite{Huang_2021_CVPR} or using nearest labeled latent vectors~\cite{Yasarla_2020_CVPR}. So, we can define error function $L$ for every image with help of similarity index($\psi$), \thatis~$L=-\psi$. Here, similarity index ($\psi$) for each image is computed as,  $\psi = \frac{1}{M_{NN}}\sum_{z^{k}\in NN(z)} \frac{\langle z,z^{k}\rangle}{|z||z^{k}|}$, where $NN(z)$ nearest neighbor of $z$ and $M_{NN}$ number of nearest neighbors. Fig~\ref{fig:psi_graph} shows sample normalized histogram graphs corresponding to $\psi$.  In Fig~\ref{fig:psi_graph}, we can observe that there is some domain gap between labeled and unlabeled images. We can also deduce the fact from Fig~\ref{fig:psi_graph} that a few unlabeled images are similar to the labeled images that will help to improve the SSR performance and a few unlabeled images may hurt the SSR performance. According to Corollary 2 to make a SSR method specified we should create subset $\mathcal{T}_1$, where unlabeled images in $\mathcal{T}_1$ should satisfy $|L_u-L_l|<\epsilon$ and $\epsilon$ is small positive number. 
	  Hence, one can come up with a rejection rule to reject the unlabeled images that might hurt the SSR performance.  To this end, we propose a novel rejection technique where we adaptively update the threshold $T$ using $\psi$ values and aelotoric uncertainity~\cite{kendall2017uncertainties}. Note, we compute aleotoric uncertainty~\cite{kendall2017uncertainties} variance $\sigma$ that makes ART-SS robust to network's errors in latent representations, since $\sigma$ indicates how confident the network is about the computed latent representation vector $z$. By re-scalinng $\psi$ values with $\sigma$, we will be giving higher importance to highly confident or less importance to less confident image samples  while computing threshold $T$ and rejecting the unnlabeled images.

	In our adaptive rejection technique, we project every labeled image $x^l_i$ and unlabeled image $x^u_i$ to a latent space and obtain $z^l_i,\sigma^l_i$ and $z^u_i,\sigma^u_i$ respectively using aleotoric uncertainity.  For more details on how to compute $\sigma$, please refer to the supplementary document. Thus, we obtain $Z^{src},\{\sigma^l\}^{src}$ and $Z^{tgt},\{\sigma^u\}^{tgt}$. 
	Having obtained $Z^{src},\{\sigma^l\}^{src}$ and $Z^{tgt},\{\sigma^u\}^{tgt}$ values, we compute $\psi^l_i$ and $\psi^u_i$ for each  labeled $x^l_i$ and unlabeled $x^u_i$ image respectively. We define the threshold, $T$, as a weighted mean $\psi^l_i$ values of the labeled images, \thatis,
	\begin{equation}
	\resizebox{0.9\hsize}{!}{$
		\Psi^{src} = \left\lbrace\psi^l_i: \psi^l_i = \frac{1}{M_{NN}}\sum_{z^{k}\in NN(z^l_i)} \frac{\langle z_i^l,z^{k}\rangle}{|z^l_i||z^{k}|}, z^l_i = g_\theta(x^l_i), \forall x^l_i \in \mathcal{D}_{src} \right\rbrace ,
		T = \frac{1}{N_\ell} \sum_{\Psi^{src},\{\sigma^l\}^{src}} \frac{\psi^l_i}{\sigma^l_i},$}
	\end{equation} 
	where $g_\theta(.)$ is the encoder of the network $f_\theta$, and we use $\sigma^l_i$ values implying higher importance is given to highly confident samples in deciding the threshold $T$. During semi-supervised training of network $f_\theta$, we will reject the unlabeled image, if $\frac{\psi^u_i}{\sigma^u_i}< T$. Fig~\ref{fig:Method} gives the overview of the proposed adaptive rejection technique. We also provide a pseudo algorithm for the proposed rejection technique in the supplementary document.
	
	Thus, following~\cite{wei2019semi,Yasarla_2020_CVPR,Huang_2021_CVPR,shao2020domain,li2019semi} we train a semi-supervised network in two phases: (i) labeled training phase, and (ii) unlabeled training phase. In the labeled training phase, we learn the network weights $f_\theta$ using the labeled images $x^l_i \in \mathcal{D}_{src}$ in a fully-supervised fashion. Additionally, we compute $\{\psi^l_i\}$ and $\{\sigma^l_i\}$ for all the labeled images,\thatis~ we compute $\Psi^{src}$ and decide threshold $T$  as explained earlier.
	In the unlabeled training phase, given an unlabeled images $x^u_i \in \mathcal{D}_{tgt}$, we compute $\{\psi^u_i\}$  and $\{\sigma^u_i\}$ values For each unlabeled image, we check the criterion: $\frac{\psi^u_i}{\sigma^u_i}< T$, and decide whether to use unlabeled image for updating the network weights $f_{\theta}$ using $\mathcal{L}_{unsup}$. Note that $\mathcal{L}_{unsup}$ can be an unsupervised loss proposed in corresponding SSR method~\cite{wei2019semi,Yasarla_2020_CVPR,Huang_2021_CVPR,Huang_2021_CVPR,shao2020domain,li2019semi}.

	\section{Experiments}
	 To show the effectiveness of ART-SS  we conduct experiments on existing SSR methods~\cite{wei2019semi,Yasarla_2020_CVPR,Huang_2021_CVPR,shao2020domain,li2019semi} showing that ART-SS boosts their performance. We conduct cross-domain experiments which cover 1) synthetic-to-synthetic experiments (where $\mathcal{D}_{src}$ and $\mathcal{D}_{tgt}$ contains synthetic rain images) and 2) synthetic-to-real (where $\mathcal{D}_{src}$ contains synthetic rain and $\mathcal{D}_{tgt}$ contains real rain).
	\vspace{-1em}
	\subsection{Datasets and Metrics}
    \noindent{\textbf{Synthetic deraining datasets.}} (i) Rain800 proposed by Zhang~\etal~\cite{zhang2019image} which contains 700 synthetic paired training images and 100 synthetic paired test images. (ii) Rain200H dataset published by Yang~\etal~\cite{yang2017deep} which contains synthetic 1,800 paired training images and 200 paired test images. (iii) Rain1400 proposed by Fu~\etal~\cite{fu2017removing} which contains 9,100 synthetic pairs for training and 1,400 pairs in the test set. (iv) Rain1200 introduced by Zhang~\etal~\cite{zhang2018density} which consists of 12,000 synthetic pairs for training, and 1,200 pairs in the test set.\\
    \noindent{\textbf{Real rainy image datasets.}}  Wang~\etal\cite{wang2019spatial} constructed a real rainy image dataset, called SPA-data, which contains paired 342 high resolution real rain frames extracted from videos for training. SPA-data contains 1,000 real rainy image pairs in the  test set.
    
    Wei~\etal~\cite{wei2019semi} created the DDN-SIRR dataset which has both labeled synthetic(9100 training images from Rain1400) and unlabeled (147 real-world rainy images) for training of semi-supervised deraining methods. Furthermore, a test set for DDN-SIRR is created using 10 dense and 10 sparse rain streak images.\\
    \noindent{\textbf{Dehazing datasets.}}Following~\cite{shao2020domain,li2019semi}, we create training source($\mathcal{D}_{src}$) and target dataset($\mathcal{D}_{tgt}$) using RESIDE\cite{li2018benchmarking}(contains ITS (Indoor Training Set), OTS (Outdoor Training Set), SOTS (Synthetic Object Testing Set), URHI (Unlabeled real Hazy Images), and RTTS (Real Task-driven Testing Set)). Labeled training set(Syn-haze) is constructed using randomly selectinng 3000 from ITS and 3000 from OTS. 2000 random images from URHI are used as unlabeled training set. \\
	\noindent{\textbf{Metrics.}} We use peak-signal-to-noise ratio (PSNR) and structural similarity index measure (SSIM) to compare the performance of different methods.
	\vspace{-1em}
	\subsection{Implementation}
	We performed our experiments on three existing SSD methods~\cite{wei2019semi,Yasarla_2020_CVPR,Huang_2021_CVPR}. We follow the same instructions and settings provided in the corresponding papers to train their SSD model in a semi-supervision fashion.\\
	\noindent{\textbf{SIRR.}} The authors use a DerainNet~\cite{fu2017removing} to perform deraining. We extract the 16th layer output and define it as the latent vector $z$. For more details about the labeled and unlabeled training phases please refer\cite{wei2019semi}. Additionally, we compute $\{z^l_i,\:\sigma^l_i,\:\psi^l_i\}$, and threshold $T$ in labeled training phase. In unlabeled training phase we model $GMM_{real}$ using the  unlabeled images that follow the criterion in Section \ref{sec:reject_tech} to while modelling $GMM_{real}$. Finally, KL divergence is used to minimize the distribution difference between $GMM_{syn}$ and $GMM_{real}$.
	\\
	\noindent{\textbf{Syn2Real.}} The authors use an encoder-decoder network constructed using Res2Block~\cite{gao2019res2net} to perform deraining. We use encoder output as the latent vector $z$. In the labeled training phase authors perform L1-norm minimization using paired images from $\mathcal{D}_{src}$.  In the unlabeled training phase, we apply the proposed ART-SS method and compute the pseudo-ground truths (pseudo-GTs)  using Gaussian process for the images which follow the criterion in Section \ref{sec:reject_tech}. Finally, we update the network weights using these computed pseudo-GTs using $\mathcal{L}_{unsup}$ Please refer to~\cite{Yasarla_2020_CVPR} for more training details.\\
	\noindent{\textbf{MOSS.}} The authors constructed a UNet-based~\cite{ronneberger2015u} deraining network using an encoder and a decoder with residual blocks. We use the encoder output as the latent vector $z$.  Obtaining the latent vectors for the labeled and unlabeled images ($z^l_i$ and $z^u_i$ respectively) we apply our rejection technique, to reject the unlabeled images that are hurting the semi-supervised performance of MOSS. For more details about the labeled and unlabeled training phases refer to~\cite{Huang_2021_CVPR}. \\
	\noindent{\textbf{Li~\etal~ and DAID}}. For Li~\etal\cite{li2019semi} we define 15th layer output as latent $z$. On the other hand, for DAID~\cite{shao2020domain} we define 12th layer output as latent $z$. Please refer\cite{li2019semi,shao2020domain} for corresponding labeled and unlabeled training phases details. Given the latent labeled and unlabeled vectors obtained from the network, we apply our ART-SS and use the unlabeled images that follow the criterion in Section \ref{sec:reject_tech} during unlabeled training phase.\\
	Note that we compute $\{z^l_i,\:\sigma^l_i,\:\psi^l_i\}$,  $\{z^u_i,\:\sigma^u_i,\:\psi^l_i\}$,  (every iteration) and threshold $T$ (every epoch) in order to apply our ART-SS to these SSR methods. 
	
	\vspace{-1.5em}
	\begin{table*}[h!]
		\caption{Quantitative gains obtained for SSD methods~\cite{wei2019semi,Yasarla_2020_CVPR,Huang_2021_CVPR} using our ART-SS method on DDN-SIRR dataset. Here synthetic labeled images of DDN-SIRR dataset are used as $\mathcal{D}_{src}$, and real rainy images as $\mathcal{D}_{tgt}$. Note gains are indicated in the brackets.}
		\label{tab:ddnsirr_synthetic}
		\centering
		\vskip-10pt
		\resizebox{1\linewidth}{!}{
			\begin{tabular}{|l|c|cccccc|ccc|ccc|ccc|}
				\hline
				\multirow{3}{*}{{Dataset}} & \multirow{3}{*}{\begin{tabular}[c]{@{}c@{}}Input\end{tabular}} & \multicolumn{6}{c|}{Methods that use only synthetic dataset}                                                                                  & \multicolumn{9}{c|}{Methods that use synthetic and real-world dataset} \\ \cline{3-17} 
				&                                                                & \multirow{2}{*}{\begin{tabular}[c]{@{}c@{}}JORDER \cite{yang2017deep}\\(CVPR '17)\end{tabular}} & \multirow{2}{*}{\begin{tabular}[c]{@{}c@{}}DDN \cite{fu2017removing}\\(CVPR '17)\end{tabular}}& \multirow{2}{*}{\begin{tabular}[c]{@{}c@{}}PReNet \cite{ren2019progressive}\\(CVPR '19)\end{tabular}} & \multirow{2}{*}{\begin{tabular}[c]{@{}c@{}}MSPFN \cite{jiang2020multi}\\(CVPR '20)\end{tabular}} &
				\multirow{2}{*}{\begin{tabular}[c]{@{}c@{}}DRD \cite{deng2020detail}\\(CVPR '20)\end{tabular}} &
				\multirow{2}{*}{\begin{tabular}[c]{@{}c@{}}MPRN \cite{zamir2021multi}\\(CVPR '21)\end{tabular}} & \multicolumn{3}{c|}{SIRR \cite{wei2019semi} (CVPR '19)}  & \multicolumn{3}{c|}{Syn2Real\cite{Yasarla_2020_CVPR}(CVPR'20)}    & \multicolumn{3}{c|}{MOSS\cite{Huang_2021_CVPR} (CVPR '21)}             \\ \cline{9-17} 
				&                                                                                              &                     &                         &                      &                      &                          &      & w/o SSD   & SSD w/o ART-SS & SSD w/ ART-SS  &  w/o SSD   & SSD w/o ART-SS & SSD w/ ART-SS  & w/o SSD   & SSD w/o ART-SS & SSD w/ ART-SS             \\ \hline
				Dense                    & 17.95                                                                                  & 18.75                   & 19.90                & 20.65     & 19.54 & 20.34 & 20.87           & 20.01   & 21.60(1.59) &  22.16({\color{blue}2.15})   & 20.24  & 22.36(2.12) &22.67({\color{blue}2.43}) & 20.29 & 22.91(2.62) & 23.32({\color{blue}3.02})   \\
				Sparse                   & 24.14                                                                                 & 24.22                   & 26.88                &  26.40 & 26.47 & 26.04 & 26.28     & 26.90   & 26.98(0.08) & 27.21({\color{blue}0.31})   & 26.15  & 27.12(0.97) & 27.48({\color{blue}1.33}) & 25.90 & 27.78(1.88) & 28.16({\color{blue}2.26})  \\ \hline
			\end{tabular}
		}
	\end{table*}
	\vspace{-3.5em}
	\begin{table*}[h!]
		\caption{PSNR/SSIM comparisons for SSR methods~\cite{li2019semi,shao2020domain} using our ART-SS method. Here synthetic labeled images of Syn-haze are used as $\mathcal{D}_{src}$, and real hazy images of URHI as $\mathcal{D}_{tgt}$.}
		\label{tab:haze_synthetic}
		\centering
		\vskip-10pt
		\resizebox{1\linewidth}{!}{
			\begin{tabular}{|l|c|ccccc|cc|cc|}
        \hline
        \multirow{2}{*}{Test set} & \multirow{2}{*}{Haze} & \multirow{2}{*}{DCP\cite{he2010single}} & \multirow{2}{*}{DehazeNet\cite{cai2016dehazenet}} & \multirow{2}{*}{DPCDN\cite{zhang2018densely}} & \multirow{2}{*}{GFN\cite{ren2018gated}} & \multirow{2}{*}{EPDN\cite{qu2019enhanced}} & \multicolumn{2}{c|}{Li\etal\cite{li2019semi}} & \multicolumn{2}{c|}{DAID\cite{shao2020domain}} \\ \cline{8-11} 
         &  &  &  &  &  &  & w/o ART-SS & w/ ART-SS & w/o ART-SS & w/ ART-SS \\ \hline
        SOTS & 13.95/0.64 & 15.49/0.64 & 21.14/0.85 & 19.39/0.65 & 22.30/0.88 & 23.82/0.89 & 24.44/0.89 & 25.56/0.92 & 27.76/0.93 & 29.15/0.95 \\ \hline
        HazeRD & 14.01/0.39 & 14.01/0.39 & 15.54/0.41 & 16.12/0.34 & 13.98/0.37 & 17.37/0.56 & 16.55/0.47 & 18.17/0.56 & 18.07/0.63 & 19.50/0.66 \\ \hline
        \end{tabular}
		}
	\end{table*}
	\vspace{-3em}
	\subsection{Comparisons}
\begin{figure*}[h!]
		\centering
		\includegraphics[width=1.0\linewidth]{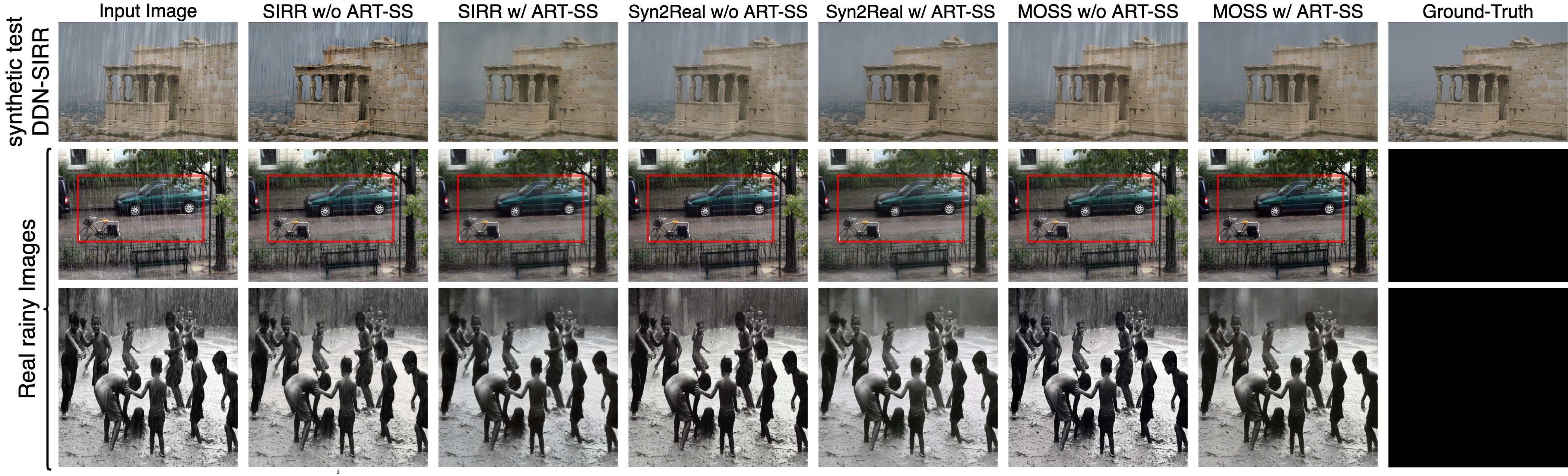} 
		\vskip -10pt 
		\caption{Qualitative comparisons showing the benefits of using ART-SS technique for SSR methods~\cite{wei2019semi,Yasarla_2020_CVPR,Huang_2021_CVPR}. First, second, third rows: $\mathcal{D}_{src} $ = synthetic labeled images from DDN-SIRR, and $\mathcal{D}_{tgt} $ = real unlabeled images from DDN-SIRR.}
		\label{fig:qual}
	\end{figure*}

\begin{figure*}[h!]
		\centering
		\includegraphics[width=1.0\linewidth]{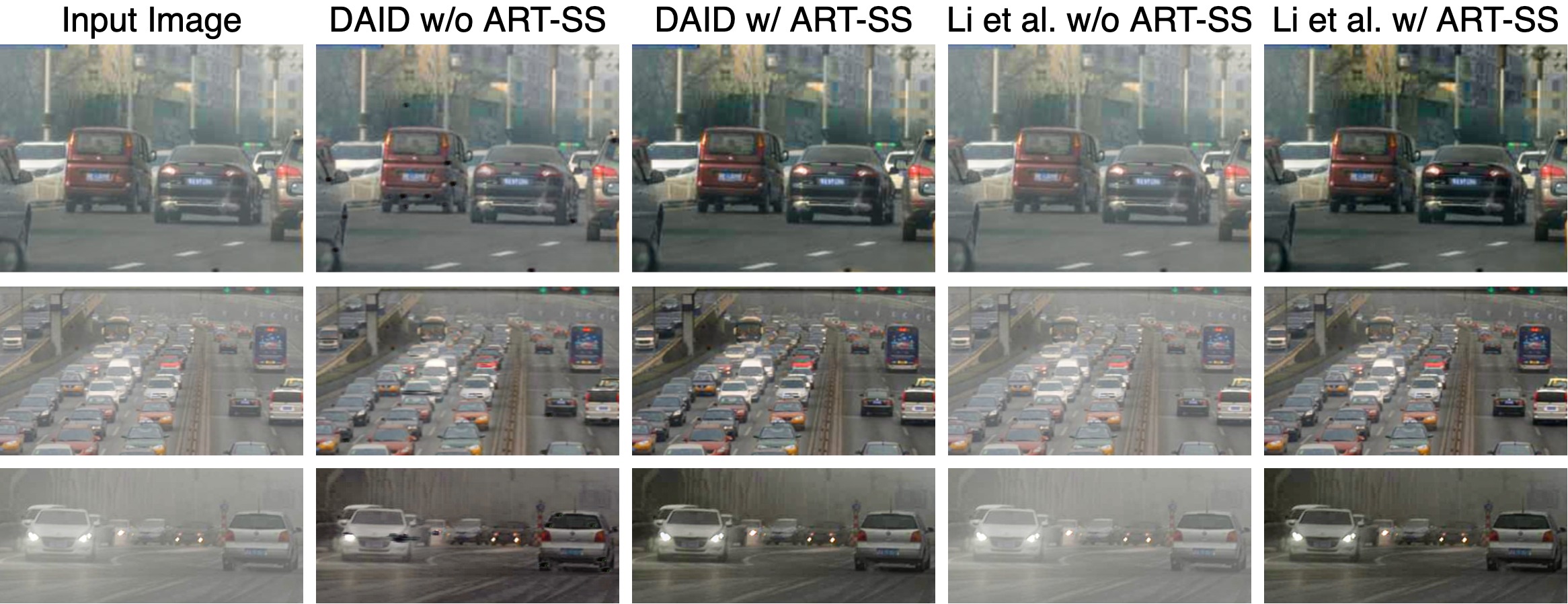} 
		\vskip -10pt 
		\caption{Qualitative comparisons showing the benefits of using ART-SS for SSR methods~\cite{li2019semi,shao2020domain} using real hazy images from RTTS.}
		\label{fig:qual_haze}
	\end{figure*}

	\noindent{\textbf{DDN-SIRR}.}  Following the protocol introduced by \cite{wei2019semi}, in this experiment, we train the SSR methods \cite{wei2019semi,Yasarla_2020_CVPR,Huang_2021_CVPR} where we set $\mathcal{D}_{src}$ as the synthetic labeled data of DDN-SIRR, and $\mathcal{D}_{tgt}$ as the real rainy image unlabeled data of DDN-SIRR. On the other hand, fully-supervised methods \cite{yang2017deep,fu2017removing,ren2019progressive,jiang2020multi,deng2020detail,zamir2021multi}  only use $\mathcal{D}_{src}$, synthetic labeled data of DDN-SIRR for training. Table~\ref{tab:ddnsirr_synthetic} shows the quantitative results on the synthetic test set of DDN-SIRR. We can observe that SSR methods~\cite{wei2019semi,Yasarla_2020_CVPR,Huang_2021_CVPR} outperform fully-supervised methods \cite{yang2017deep,fu2017removing,ren2019progressive,jiang2020multi,deng2020detail,zamir2021multi}, since they leverage information from unlabeled images in $\mathcal{D}_{tgt}$ during training. However, there is still a room for improving the performance of these SSR methods~\cite{wei2019semi,Yasarla_2020_CVPR,Huang_2021_CVPR}.   As can be seen from Table~\ref{tab:ddnsirr_synthetic}, when we use the proposed rejection method to reject samples from the unlabeled target domain, we observe a significant improvement in the performance of these SSR methods.  Results are shown in Table~\ref{tab:ddnsirr_synthetic} . 
	We also provide qualitative results on one example from the synthetic test set of DDN-SIRR, and two real rain examples,  in Fig~\ref{fig:qual}. As can be seen from this figure, the output images \cite{wei2019semi,Yasarla_2020_CVPR,Huang_2021_CVPR} without ART-SS still contain some rain streaks and are of low-quality (see the highlighted red box where the network under-performed). On the other hand, \cite{wei2019semi,Yasarla_2020_CVPR,Huang_2021_CVPR} with ART-SS achieve
	better quality derained output images.\\
	\noindent{\textbf{De-haze experiments.}} Following the protocol introduced in~\cite{li2019semi,shao2020domain}, we train SSR methods\cite{li2019semi,shao2020domain} using Syn-Haze (as labeled $\mathcal{D}_{src}$) and URHI(as unlabeled $\mathcal{D}_{tgt}$). We use SOTS and HazeRD test sets for comparing SSR methods~\cite{li2019semi,shao2020domain} performance. Table~\ref{tab:haze_synthetic} shows the proposed ART-SS improved the SSR~\cite{li2019semi,shao2020domain} performance by around 1.4dB in PSNR. Fig~\ref{fig:qual_haze} shows the qualitative comparisons on real haze images from RTTS, we can see visual quality of dehazed images by SSR~\cite{li2019semi,shao2020domain} improved when trained with ART-SS .\\
	\vspace{-2.5em}
	\vspace{-4em}
	\vspace{-2em}
	
	\noindent{\textbf{Cross-domain experiments.}} In Table~\ref{tab:cross_exp1}, we set $\mathcal{D}_{src}$ as Rain800, and  $\mathcal{D}_{tgt} $ as Rain1400, Rain200L, and SPA-data to train the SSD methods \cite{wei2019semi,Yasarla_2020_CVPR,Huang_2021_CVPR}. In Table~\ref{tab:cross_exp2}, we set $\mathcal{D}_{src} $ as Rain1400, and  $\mathcal{D}_{tgt} $ as Rain800, Rain200L, and SPA-data.
	From Table~\ref{tab:cross_exp2}, and Table~\ref{tab:cross_exp1}, we can clearly see that SSR methods \cite{wei2019semi,Yasarla_2020_CVPR,Huang_2021_CVPR} with ART-SS outperform methods without rejection with huge margin in PSNR and SSIM. Furthermore, from Table~\ref{tab:cross_exp2}, and Table~\ref{tab:cross_exp1}, it is evident that ART-SS is beneficial in improving the performance of \cite{wei2019semi,Yasarla_2020_CVPR,Huang_2021_CVPR}, even in variations in the dataset sizes of $\mathcal{D}_{src}$ and $\mathcal{D}_{tgt}$, and different synthetic and real rain datasets. 
	\vspace{-2em}
	\begin{table*}[h!]
		\caption{Cross-domain experiment with $\mathcal{D}_{src}$ as  Rain1400, and $\mathcal{D}_{tgt}$ as Rain800, Rain200L and SPA-data.  Note gains are indicated in the brackets. We highlight gains obtained using our ART-SS rejection technique for SSD methods~\cite{wei2019semi,Yasarla_2020_CVPR,Huang_2021_CVPR} with {\color{blue}blue}.}
		\label{tab:cross_exp2}
		\centering
		\vskip-10pt
		\resizebox{1.0\hsize}{!}{
			\begin{tabular}{|l|l|cccccc|llllll|llllll|}
				\hline
				\multirow{3}{*}{\begin{tabular}[c]{@{}l@{}}Source dataset\\ $\mathcal{D}_{src}$\end{tabular}} & SSD method & \multicolumn{6}{c|}{SIRR} & \multicolumn{6}{c|}{Syn2Real} & \multicolumn{6}{c|}{MOSS} \\ \cline{2-20} 
				& \multirow{2}{*}{\begin{tabular}[c]{@{}l@{}}Target dataset\\ $\mathcal{D}_{tgt}$\end{tabular}} & \multicolumn{2}{c}{Rain800} & \multicolumn{2}{c}{Rain200L} & \multicolumn{2}{c|}{SPA-data} & \multicolumn{2}{c}{Rain800} & \multicolumn{2}{c}{Rain200L} & \multicolumn{2}{c|}{SPA-data} & \multicolumn{2}{c}{Rain800} & \multicolumn{2}{c}{Rain200L} & \multicolumn{2}{c|}{SPA-data} \\
				&  & PSNR & SSIM & PSNR & SSIM & PSNR & SSIM & \multicolumn{1}{c}{PSNR} & \multicolumn{1}{c}{SSIM} & \multicolumn{1}{c}{PSNR} & \multicolumn{1}{c}{SSIM} & \multicolumn{1}{c}{PSNR} & \multicolumn{1}{c|}{SSIM} & \multicolumn{1}{c}{PSNR} & \multicolumn{1}{c}{SSIM} & \multicolumn{1}{c}{PSNR} & \multicolumn{1}{c}{SSIM} & \multicolumn{1}{c}{PSNR} & \multicolumn{1}{c|}{SSIM} \\ \hline
				\multirow{3}{*}{$\mathcal{D}_{src}$ = Rain1400} & \begin{tabular}[c]{@{}l@{}}Source only $\mathcal{D}_{src}$\\ w/o SSD\end{tabular} & 22.17 & 0.828 & 24.82 & 0.867 & 33.28 & 0.941 & 22.59 & 0.845 & 27.08 & 0.907 & 31.58 & 0.955 & 21.80 & 0.824 & 25.77 & 0.881 & 32.01 & 0.948 \\ \cline{2-20} 
				& \begin{tabular}[c]{@{}l@{}}SSD w/o ART-SS\\ $\mathcal{D}_{src}+\mathcal{D}_{tgt}$\end{tabular} & \begin{tabular}[c]{@{}c@{}} 22.62\\ (0.45)\end{tabular}& \begin{tabular}[c]{@{}c@{}} 0.833\\ (0.005)\end{tabular}& \begin{tabular}[c]{@{}c@{}} 26.75\\ (1.93)\end{tabular}& \begin{tabular}[c]{@{}c@{}} 0.882\\ (0.015)\end{tabular}& \begin{tabular}[c]{@{}c@{}} 34.07\\ (0.79)\end{tabular}& \begin{tabular}[c]{@{}c@{}} 0.950\\ (0.009)\end{tabular} & \begin{tabular}[c]{@{}c@{}} 22.87\\ (0.28)\end{tabular} & \begin{tabular}[c]{@{}c@{}} 0.846\\ (0.001)\end{tabular}& \begin{tabular}[c]{@{}c@{}} 27.97\\ (0.89)\end{tabular}& \begin{tabular}[c]{@{}c@{}} 0.929\\ (0.022)\end{tabular}& \begin{tabular}[c]{@{}c@{}} 34.24\\ (2.66)\end{tabular} & \begin{tabular}[c]{@{}c@{}} 0.962\\ (0.007)\end{tabular}&
				\begin{tabular}[c]{@{}c@{}} 22.56\\ (0.76)\end{tabular}& \begin{tabular}[c]{@{}c@{}} 0.831\\ (0.007)\end{tabular} & 
				\begin{tabular}[c]{@{}c@{}} 27.09\\ (1.32)\end{tabular}& \begin{tabular}[c]{@{}c@{}} 0.896\\  (0.015)\end{tabular}&\begin{tabular}[c]{@{}c@{}} 34.56\\ (2.55)\end{tabular} & \begin{tabular}[c]{@{}c@{}} 0.960\\ (0.012)\end{tabular}\\ \cline{2-20} 
				& \begin{tabular}[c]{@{}l@{}}SSD w/ ART-SS\\ $\mathcal{D}_{src}+\mathcal{D}_{tgt}$\end{tabular} & \begin{tabular}[c]{@{}c@{}}23.42\\ ({\color{blue}1.25})\end{tabular}  & \begin{tabular}[c]{@{}c@{}} 0.841\\ ({\color{blue} 0.012})\end{tabular}  & \begin{tabular}[c]{@{}c@{}} 29.38\\ ({\color{blue}4.56})\end{tabular} & \begin{tabular}[c]{@{}c@{}} 0.910\\ ({\color{blue} 0.043})\end{tabular} & \begin{tabular}[c]{@{}c@{}} 36.12\\ ({\color{blue} 2.84})\end{tabular} & \begin{tabular}[c]{@{}c@{}} 0.962\\ ({\color{blue} 0.021})\end{tabular}  & \begin{tabular}[c]{@{}c@{}} 23.85\\ ({\color{blue} 1.26})\end{tabular} & \begin{tabular}[c]{@{}c@{}} 0.863\\ ({\color{blue} 0.018})\end{tabular}& \begin{tabular}[c]{@{}c@{}} 30.76\\ ({\color{blue} 3.68})\end{tabular} & \begin{tabular}[c]{@{}c@{}} 0.948\\ ({\color{blue} 0.041})\end{tabular} & \begin{tabular}[c]{@{}c@{}} 37.36\\ ({\color{blue} 5.78})\end{tabular} & \begin{tabular}[c]{@{}c@{}} 0.974\\ ({\color{blue} 0.019})\end{tabular} & \begin{tabular}[c]{@{}c@{}} 24.32\\ ({\color{blue} 2.52})\end{tabular} &
				\begin{tabular}[c]{@{}c@{}} 0.870\\ ({\color{blue} 0.046})\end{tabular}  & \begin{tabular}[c]{@{}c@{}} 30.42\\ ({\color{blue} 4.65})\end{tabular}  & \begin{tabular}[c]{@{}c@{}} 0.939\\ ({\color{blue} 0.059})\end{tabular} & \begin{tabular}[c]{@{}c@{}} 37.94\\ ({\color{blue} 5.93})\end{tabular} & \begin{tabular}[c]{@{}c@{}} 0.983\\ ({\color{blue} 0.035})\end{tabular}\\ \hline
			\end{tabular}
		}
	\end{table*}
	\vspace{-4em}
	\begin{table*}[h!]
		\caption{Cross-domain experiment with $\mathcal{D}_{src}$ as Rain800, and $\mathcal{D}_{tgt}$ as Rain1400, Rain200L and SPA-data. We highlight gains obtained using our ART-SS rejection technique for SSD methods~\cite{wei2019semi,Yasarla_2020_CVPR,Huang_2021_CVPR} with {\color{blue}blue}.}
		\label{tab:cross_exp1}
		\centering
		\vskip-10pt
		\resizebox{1.0\hsize}{!}{
			\begin{tabular}{|l|l|cccccc|llllll|llllll|}
				\hline
				\multirow{3}{*}{\begin{tabular}[c]{@{}l@{}}Source dataset\\ $\mathcal{D}_{src}$\end{tabular}} & SSD method & \multicolumn{6}{c|}{SIRR} & \multicolumn{6}{c|}{Syn2Real} & \multicolumn{6}{c|}{MOSS} \\ \cline{2-20} 
				& \multirow{2}{*}{\begin{tabular}[c]{@{}l@{}}Target dataset\\ $\mathcal{D}_{tgt}$\end{tabular}} & \multicolumn{2}{c}{Rain1400} & \multicolumn{2}{c}{Rain200L} & \multicolumn{2}{c|}{SPA-data} & \multicolumn{2}{c}{Rain1400} & \multicolumn{2}{c}{Rain200L} & \multicolumn{2}{c|}{SPA-data} & \multicolumn{2}{c}{Rain1400} & \multicolumn{2}{c}{Rain200L} & \multicolumn{2}{c|}{SPA-data} \\
				&  & PSNR & SSIM & PSNR & SSIM & PSNR & SSIM & \multicolumn{1}{c}{PSNR} & \multicolumn{1}{c}{SSIM} & \multicolumn{1}{c}{PSNR} & \multicolumn{1}{c}{SSIM} & \multicolumn{1}{c}{PSNR} & \multicolumn{1}{c|}{SSIM} & \multicolumn{1}{c}{PSNR} & \multicolumn{1}{c}{SSIM} & \multicolumn{1}{c}{PSNR} & \multicolumn{1}{c}{SSIM} & \multicolumn{1}{c}{PSNR} & \multicolumn{1}{c|}{SSIM} \\ \hline
				\multirow{3}{*}{$\mathcal{D}_{src}$ = Rain800} & \begin{tabular}[c]{@{}l@{}}Source only $\mathcal{D}_{src}$\\ w/o SSD\end{tabular} & 24.64 & 0.871 & 24.78 & 0.881 & 31.75 & 0.937 & 25.17 & 0.903 & 27.02 & 0.923 & 31.36 & 0.959 & 24.98 & 0.888 & 26.75 & 0.923 & 32.09 & 0.946 \\ \cline{2-20} 
				& \begin{tabular}[c]{@{}l@{}}SSD w/o ART-SS\\ $\mathcal{D}_{src}+\mathcal{D}_{tgt}$\end{tabular} & \begin{tabular}[c]{@{}c@{}} 26.17\\ (1.53)\end{tabular}& \begin{tabular}[c]{@{}c@{}} 0.889\\ (0.018)\end{tabular}& \begin{tabular}[c]{@{}c@{}} 26.56\\ (1.78)\end{tabular}& \begin{tabular}[c]{@{}c@{}} 0.897\\ (0.016)\end{tabular}& \begin{tabular}[c]{@{}c@{}} 33.89\\ (2.14)\end{tabular}& \begin{tabular}[c]{@{}c@{}} 0.946\\ (0.009)\end{tabular} & \begin{tabular}[c]{@{}c@{}} 26.38\\ (1.21)\end{tabular} & \begin{tabular}[c]{@{}c@{}} 0.911\\ (0.008)\end{tabular}& \begin{tabular}[c]{@{}c@{}} 27.65\\ (0.63)\end{tabular}& \begin{tabular}[c]{@{}c@{}} 0.930\\ (0.007)\end{tabular}& \begin{tabular}[c]{@{}c@{}} 34.16\\ (2.80)\end{tabular} & \begin{tabular}[c]{@{}c@{}} 0.966\\ (0.007)\end{tabular}&
				\begin{tabular}[c]{@{}c@{}} 26.84\\ (1.86)\end{tabular}& \begin{tabular}[c]{@{}c@{}} 0.904\\ (0.016)\end{tabular} & 
				\begin{tabular}[c]{@{}c@{}} 27.67\\ (0.92)\end{tabular}& \begin{tabular}[c]{@{}c@{}} 0.928\\  (0.005)\end{tabular}&\begin{tabular}[c]{@{}c@{}} 34.82\\ (2.73)\end{tabular} & \begin{tabular}[c]{@{}c@{}} 0.962\\ (0.016)\end{tabular}\\ \cline{2-20} 
				& \begin{tabular}[c]{@{}l@{}}SSD w/ ART-SS\\ $\mathcal{D}_{src}+\mathcal{D}_{tgt}$\end{tabular} & \begin{tabular}[c]{@{}c@{}}26.98\\ ({\color{blue}2.34})\end{tabular}  & \begin{tabular}[c]{@{}c@{}} 0.903\\ ({\color{blue} 0.032})\end{tabular}  & \begin{tabular}[c]{@{}c@{}} 29.10\\ ({\color{blue}4.32})\end{tabular} & \begin{tabular}[c]{@{}c@{}} 0.918\\ ({\color{blue} 0.037})\end{tabular} & \begin{tabular}[c]{@{}c@{}} 36.32\\ ({\color{blue} 4.57})\end{tabular} & \begin{tabular}[c]{@{}c@{}} 0.962\\ ({\color{blue} 0.025})\end{tabular}  & \begin{tabular}[c]{@{}c@{}} 27.84\\ ({\color{blue} 2.67})\end{tabular} & \begin{tabular}[c]{@{}c@{}} 0.922\\ ({\color{blue} 0.019})\end{tabular}& \begin{tabular}[c]{@{}c@{}} 30.41\\ ({\color{blue} 3.39})\end{tabular} & \begin{tabular}[c]{@{}c@{}} 0.944\\ ({\color{blue} 0.021})\end{tabular} & \begin{tabular}[c]{@{}c@{}} 37.28\\ ({\color{blue} 5.92})\end{tabular} & \begin{tabular}[c]{@{}c@{}} 0.981\\ ({\color{blue} 0.022})\end{tabular} & \begin{tabular}[c]{@{}c@{}} 29.01\\ ({\color{blue} 4.03})\end{tabular} &
				\begin{tabular}[c]{@{}c@{}} 0.919\\ ({\color{blue} 0.031})\end{tabular}  & \begin{tabular}[c]{@{}c@{}} 31.02\\ ({\color{blue} 4.27})\end{tabular}  & \begin{tabular}[c]{@{}c@{}} 0.951\\ ({\color{blue} 0.028})\end{tabular} & \begin{tabular}[c]{@{}c@{}} 37.56\\ ({\color{blue} 5.47})\end{tabular} & \begin{tabular}[c]{@{}c@{}} 0.982\\ ({\color{blue} 0.036})\end{tabular}\\ \hline
				  
			\end{tabular}
			}
	\end{table*}
	\vspace{-4em}
	\begin{table*}[htp!]
		\caption{Ablation study for ART-SS. Note, ``RS" mean random sampling, ``NR" means no rejection \thatis using all unlabeled images from $\mathcal{D}_{tgt}$.}
		\label{tab:abl}
		\centering
		\vskip-10pt
		\resizebox{1.0\hsize}{!}{
			\begin{tabular}{|l|l|l|c|c|c|c|c|c|c|c|c|c|}
				\hline
				\multirow{2}{*}{\begin{tabular}[c]{@{}l@{}}Source \\ dataset\end{tabular}} & \multirow{2}{*}{\begin{tabular}[c]{@{}l@{}}Target\\ dataset\end{tabular}} & \multirow{2}{*}{Metrics} & \multicolumn{5}{c|}{Syn2Real} & \multicolumn{5}{c|}{MOSS} \\ \cline{4-13} 
				&  &  & w/o SSD & SSD w/ NR & SSD w/ RS & SSD w/ $\psi$ & SSD w/ ART-SS & \multicolumn{1}{c|}{w/o SSD} &  SSD w/ NR & \multicolumn{1}{c|}{SSD w/ RS} & \multicolumn{1}{c|}{SSD w/ $\psi$} & \multicolumn{1}{c|}{SSD w/ ART-SS} \\ \hline
				\multirow{2}{*}{$\mathcal{D}_{src}$ = Rain800} & \multirow{2}{*}{$\mathcal{D}_{tgt}$ = SPA-data} & PSNR & 31.36 & 34.16 & 34.94 & 35.99 & 37.28 & 32.09 & 34.82& 35.38 & 35.86 & 37.56 \\
				&  & SSIM & 0.959 & 0.966 & 0.970 & 0.973 & 0.981 & 0.946 & 0.962 & 0.968 & 0.970 & 0.982 \\ \hline
			\end{tabular}
		}
	\vspace{-4em}
	\end{table*}
	\subsection{Ablation Study}
	We perform an ablation study to show the improvements of ART-SS over random-sampling or nearest neighbors. In this experiment, we train ~\cite{Yasarla_2020_CVPR,Huang_2021_CVPR} with Rain800 as $\mathcal{D}_{src}$ and  SPA-data as $\mathcal{D}_{tgt}$, in five different settings, (i) without semi-supervision (\thatis training with only $\mathcal{D}_{src}$), w/o SSD, (ii) with semi-supervision using all the images from  $\mathcal{D}_{tgt}$ and $\mathcal{D}_{src}$, SSD w/ NR, (iii) semi-supervision using all the images from $\mathcal{D}_{src}$ but randomly sampling $N_{\mathcal{T}}$ images from $\mathcal{D}_{tgt}$ for training, SSD w/ RS, (iv) semi-supervision using all the images from $\mathcal{D}_{src}$, and rejecting unlabeled images using the just  similarity index $\psi$ between unlabeled image and nearest neighbors, (v) semi-supervision using ART-SS, \thatis~ computing $T$ using $\psi$ and $\sigma$ values, and rejecting unlabeled image $x^u_i$ from $\mathcal{D}_{tgt}$ using $T$ and corresponding $\{\psi^u_i,\sigma^u_i\}$ values, SSD w/ ART-SS. Ablation experiment results are shown in Table~\ref{tab:abl}.  From these results, we can notice  that SSR w/ ART-SS produces significant improvements for both SSR methods ~\cite{Yasarla_2020_CVPR,Huang_2021_CVPR}, when compared to other rejection techniques.

	\section{Conclusion}
	We theoretically study the  effect of unlabeled weather-degraded observations on semi-supervised performance, and develop a novel technique called ART-SS, that rejects the unlabeled images which are not beneficial for improving the semi-supervised performance. We conduct extensive cross-domain experiments on different datasets to  show the effectiveness of proposed ART-SS technique  in improving the performance of \cite{wei2019semi,Yasarla_2020_CVPR,Huang_2021_CVPR,li2019semi,shao2020domain}.

\clearpage
%
%
\bibliographystyle{splncs04}
\bibliography{egbib}
\end{document}